\documentclass[twoside,11pt]{article}
\usepackage[x11names]{xcolor}
\usepackage{blindtext}
\usepackage{float}
\usepackage[
singlelinecheck=false 
]{caption}
\usepackage{jmlr2e}
\usepackage{amsthm,amsmath,amsfonts,amssymb}
\usepackage{algorithm} 
\usepackage[noend]{algpseudocode}
\usepackage{todonotes}
\usepackage{bbold}
\usepackage{mathtools}
\usepackage{xpatch}
\usepackage[symbol]{footmisc}

\usepackage{lastpage}
\usepackage{wrapfig,lipsum,booktabs}
\definecolor{aquamarine}{rgb}{0.5, 1.0, 0.83}
\definecolor{celadon}{rgb}{0.67, 0.88, 0.69}
\definecolor{electriclime}{rgb}{0.8, 1.0, 0.0}
\definecolor{grannysmithapple}{rgb}{0.66, 0.89, 0.63}
\definecolor{languidlavender}{rgb}{0.84, 0.79, 0.87}

\ShortHeadings{}{}
\firstpageno{1}

\usepackage{amssymb}
\usepackage{pifont}
\newcommand{\xmark}{\ding{55}}%
\usepackage{tikz}

\theoremstyle{remark}

\ifdefined\proof
    \renewenvironment{proof}{\par\noindent{\bf  }}{\hfill\BlackBox\\[2mm]}
\else
    \newenvironment{proof}{\par\noindent{\bf  }}{\hfill\BlackBox\\[2mm]}
\fi

\begin{document}
\setlength{\marginparwidth}{2cm}
\title{Identifying and Extracting Rare Disease Phenotypes with Large Language Models}

\author{\name Cathy Shyr$^*$ \email cathy.shyr@vumc.org \\
       \addr Department of Biomedical Informatics\\
       Vanderbilt University Medical Center, Nashville, TN, USA\\
       \AND
       \name Yan Hu \email yan.hu@uth.tmc.edu\\
       \addr School of Biomedical Informatics \\
       University of Texas Health Science at Houston, TX, USA\\
       \AND
       \name Paul Harris \email paul.a.harris@vumc.org\\
       \addr Department of Biomedical Informatics, Biostatistics, and \\Biomedical Engineering \\
       Vanderbilt University Medical Center, Nashville, TN, USA\\
       \AND 
        \name Hua Xu \email hua.xu@yale.edu\\
       \addr Section of Biomedical Informatics and Data Science\\
       School of Medicine, Yale University,
New Haven, CT, USA\\}
\footnotetext{*Corresponding author. Post Address: 2525 West End Avenue, Nashville, TN, 37203}
\maketitle
\vspace{-3.8em}
\begin{abstract}
Rare diseases (RDs) are collectively common and affect 300 million people worldwide. Accurate phenotyping is critical for informing diagnosis and treatment, but RD phenotypes are often embedded in unstructured text and time-consuming to extract manually. While natural language processing (NLP) models can perform named entity recognition (NER) to automate extraction, a major bottleneck is the development of a large, annotated corpus for model training. Recently, prompt learning emerged as an NLP paradigm that can lead to more generalizable results without any (zero-shot) or few labeled samples (few-shot). Despite growing interest in ChatGPT, a revolutionary large language model capable of following complex human prompts and generating high-quality responses, none have studied its NER performance for RDs in the zero- and few-shot settings. To this end, we engineered novel prompts aimed at extracting RD phenotypes and, to the best of our knowledge, are the first the establish a benchmark for evaluating ChatGPT's performance in these settings. We compared its performance to the traditional fine-tuning approach and conducted an in-depth error analysis. Overall, fine-tuning BioClinicalBERT resulted in higher performance (F1 of 0.689) than ChatGPT (F1 of 0.472 and 0.591 in the zero- and few-shot settings, respectively). Despite this, ChatGPT achieved similar or higher accuracy for certain entities (i.e., rare diseases and signs) in the one-shot setting (F1 of 0.776 and 0.725). This suggests that with appropriate prompt engineering, ChatGPT has the potential to match or outperform fine-tuned language models for certain entity types with just one labeled sample. While the proliferation of large language models may provide opportunities for supporting RD diagnosis and treatment, researchers and clinicians should critically evaluate model outputs and be well-informed of their limitations.
\end{abstract}

\begin{keywords}
  natural language processing, prompt learning, rare disease, artificial intelligence, ChatGPT
\end{keywords}

\clearpage
\section{Introduction}
Rare diseases are chronically
debilitating, often life-limiting conditions that affect 300 million individuals worldwide \citep{nguengang2020estimating}. Though individually rare (defined as affecting $<200,000$ individuals in the United States), rare diseases are collectively common and
represent a serious public health concern \citep{chung2022rare}. Because of the lack of knowledge and effective treatment options for rare diseases, patients undergo diagnostic and therapeutic odysseys, where they are diagnosed with delay and face difficulty searching for effective therapies \citep{childerhose2021therapeutic, insights2020barriers}. Rare disease odysseys have devastating medical, psychosocial, and economic consequences for patients and families, resulting in irreversible disease progression, physical suffering, emotional turmoil, and ongoing high medical costs \citep{cohen2010quality, carmichael2015going, yang2022national}. Thus, there is an urgent need to shorten rare disease odysseys, and reaching this goal requires effective diagnostic and treatment strategies.

Phenotyping is crucial for informing both strategies. Ongoing initiatives like the National Institutes of Health's Undiagnosed Diseases Network rely on deep phenotyping to generate candidate diseases for diagnosis, identify additional patients with similar clinical manifestations, and personalize treatment or disease management strategies  \citep{ tifft2014national, macnamara2019undiagnosed}. In addition, phenotyping can facilitate cohort identification and recruitment for clinical trials critical to the development of novel treatment regimes \citep{ahmad2020computable, chapman2021using}. Because of scarce nosological guidelines, however, rare diseases and their associated phenotypes are seldom represented in international classifications as structured data \citep{rath2012representation}. Instead, they are often embedded in unstructured text and require manual extraction by highly trained experts, which is laborious, costly,
and susceptible to bias depending on the clinician’s background and training. A promising alternative is to leverage natural language processing (NLP) models, which can automatically identify and extract rare disease entities, reduce manual workload, and improve phenotyping efficiency.

Automatic recognition of disease entities, or named entity recognition (NER), is an NLP task that involves the identification and categorization of disease information from unstructured text. This task is especially challenging due to the diversity, complexity, and ambiguity of rare diseases and their phenotypes, which can have different synonyms (e.g., cystic fibrosis and mucoviscidosis), abbreviations (e.g., CF for cystic fibrosis), and modifiers such as body location (e.g., small holes in front of the ear) and severity (e.g., extreme nearsightedness). Descriptions of rare disease phenotypes that are discontinuous, nested, or overlapping present additional challenges; moreover, those that range from short phrases in layman's terms (e.g., distention of the kidney) to medical jargon (e.g., hydronephrosis) may further complicate NER.

Over the last few decades, rapid evolution of NLP models led to significant advancements in NER. Early approaches relied on rules derived from extensive manual analysis \citep{wang2018clinical}; these were later superseded by sequence labeling models, including conditional random fields and recurrent neural networks, that capture contextual information between adjacent words \citep{li2015biomedical, patil2020named}. Over the last several years, the NER paradigm shifted toward transformer-based language models like BERT (Bidirectional Encoder Representations from Transformers), which achieved state-of-the-art performance on benchmark datasets \citep{vaswani2017attention, devlin2018bert}. Despite their success, a major bottleneck of training models for rare diseases or biomedical applications in general is the development of large, annotated corpora, which is a laborious process that requires manual annotation by domain experts. Recently, OpenAI released ChatGPT, a revolutionary, GPT-based (Generative Pre-trained Transformer) language model capable of following complex human prompts and generating high-quality responses without any annotated data (zero-shot) or with just a few examples (few-shot) \citep{ChatGPT, agrawal2022large, hu2023zero, chen2023large}. This capability, which provides opportunities to significantly reduce the manual burden of annotation without sacrificing model performance, is especially attractive for NER in the context of rare diseases.

Despite the proliferation of studies on biomedical NER, few have explored this topic for rare diseases. \cite{davis2013automated} and \cite{lo2021improving} developed NLP algorithms using the Unified Medical Language System Metathesaurus to recognize phenotypes for multiple sclerosis and Dravet syndrome, respectively. \cite{nigwekar2014quantifying} used an unnamed NLP software to identify patients with the terms ``calciphylaxis" or ``calcific uremic arteriolopathy" in their medical records. Recently, \cite{fabregat2018deep} and \cite{segura2022exploring} leveraged deep learning techniques, including Bidirectional Long Short Term Memory networks and BERT-based models, to recognize rare diseases and their clinical manifestations from texts. While some explored the potential of ChatGPT for diagnosing rare diseases with human-provided suggestions \citep{lee2023ai, mehnen2023chatgpt}, none have studied its performance for NER in the zero- or few-shot settings. 

To this end, our study makes the following contributions. 1) We designed new prompts for ChatGPT to extract rare diseases and their phenotypes (i.e., diseases, symptoms, and signs) in the zero- and few-shot settings. 2) To the best of our knowledge, this work is the first to establish a benchmark for evaluating ChatGPT's NER performance on a high-quality corpus of annotated texts on rare diseases \citep{martinez2022raredis}. In addition, we compared prompt learning to fine-tuning by training and evaluating a domain-specific BERT-based model on the annotated corpus. 3) We conducted an in-depth error analysis to elucidate the models' performance and 4) provided suggestions to help guide future work on NER for rare diseases.

\section{Methods}
\subsection{Dataset}
\begin{wrapfigure}{R}{0.6\textwidth}
  \begin{center}
    \includegraphics[width=0.6\textwidth]{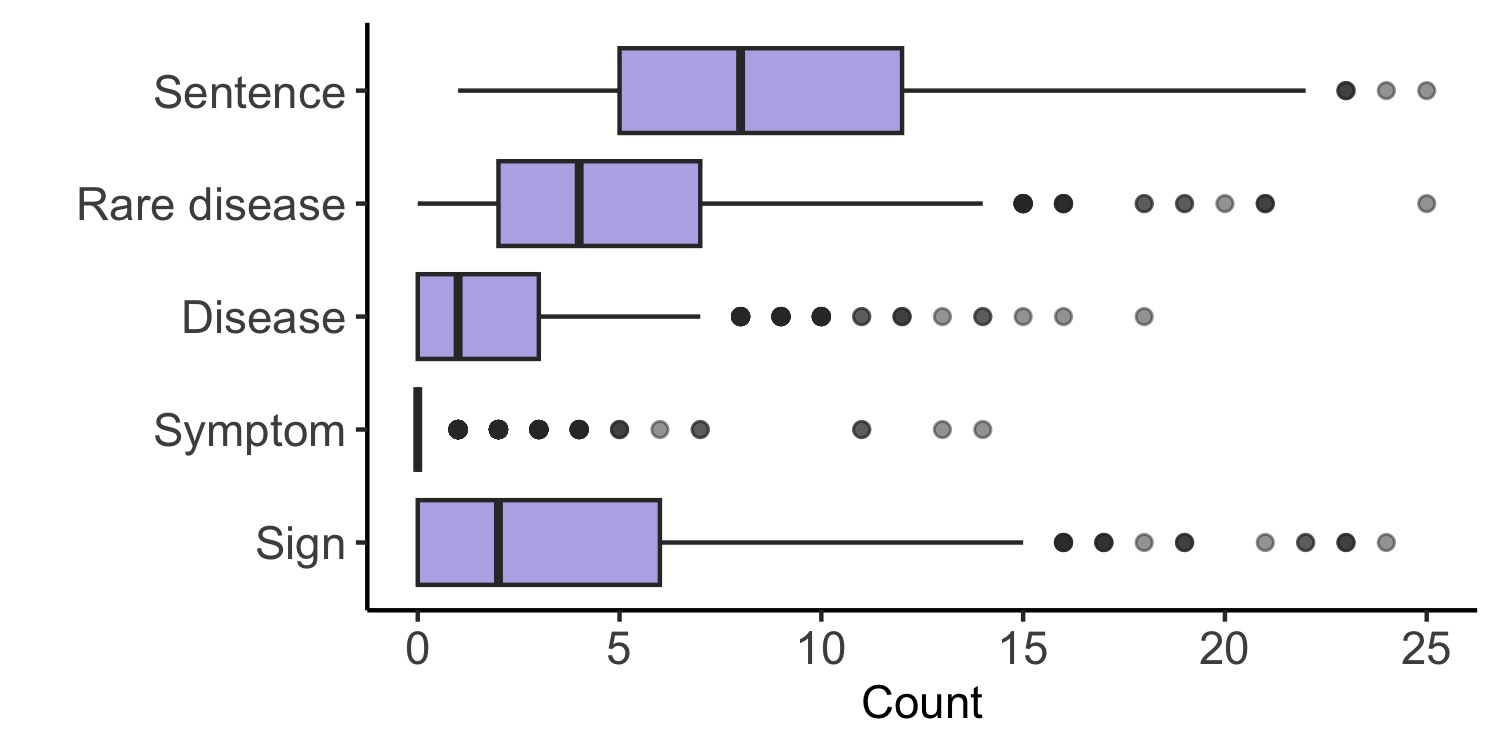}
  \end{center}
  \vspace{-2em}
  \caption{Number of sentences and entities per document.} \label{fig:count}
   \vspace{-3em}
\end{wrapfigure}

We used the RareDis corpus, which consists of $n = 832$ texts from the National Organization for Rare Disorders database \citep{martinez2022raredis}. This corpus was annotated with four entities, rare diseases, diseases, signs, and symptoms, with an inter-annotator agreement of 83.5\% under exact match. Table~\ref{tab:1} provides the entity definitions. Unlike corpora with distinct entity types, e.g., \{person, location, organization\} or \{problem, test, treatment\}, RareDis consists of entities with considerable semantic overlap. Specifically, rare diseases are a subset of diseases. Diseases can cause or be associated with other diseases as a symptom or sign. The distinction between symptoms and signs is very subtle; while both are abnormalities that may indicate a disease, the former are subjective to the patient and cannot be measured by tests or observed by physicians (e.g., pain or loss of appetite). On the other hand, a sign can be measured or observed (e.g., high blood pressure, poor lung function). Across $n = 832$ texts, there were a total of 7,354 sentences, 4,065 rare diseases, 1,814 diseases, 316 symptoms, and 3,317 signs. Rare diseases and signs were more common than diseases and symptoms, accounting for 77\% of all entities in the corpus.  Fig.~\ref{fig:count} provides a summary of counts per text. 
\begin{table}
\resizebox{\textwidth}{!}{
\begin{tabular}{lll}
\hline
\hline
Entity       & Definition    & Examples                          \\
\hline
Rare disease & Diseases which affect a small number of people                                                                                                    & cat eye syndrome, \\
& compared to the general population & Marfan syndrome\\
Disease      & An abnormal condition of a part,
organ, or system  & cancer, cardiovascular disease \\
& of an organism
resulting from various causes, such \\
& as infection, inflammation,
environmental factors, \\
& or genetic
defect, and characterized by an
identifiable \\
& group of signs,
symptoms, or both &    \\
Symptom      & A physical or mental problem that may indicate &fatigue, pain\\
& a disease or condition;
cannot be seen and do not \\
& show up
on medical tests                                                                                               &                    \\
Sign         & A physical or mental problem that may indicate & rash, abnormal heart rate         \\
& a disease or condition;
can be seen and  shows up\\
& on medical tests     \\
\hline
\end{tabular}}
\caption{Summary of entity definitions.} \label{tab:1}
\end{table}

\noindent The RareDis corpus is publicly available and distributed in the Brat standoff format \citep{stenetorp2012brat}. We refer readers to \cite{martinez2022raredis} for details on the annotation guidelines.

\subsection{NER Paradigms and Models}
We considered two popular NER paradigms for comparison: 1) pre-training + fine-tuning, and 2) prompt learning \citep{radford2018improving, liu2023pre}. The former involves a two-step process where a language model (e.g., BERT) is first trained on a massive amount of unlabeled text data and then fine-tuned on specific downstream NER tasks with labeled data. In the case of BERT models, the objective is to learn general language presentations through masked language modeling during the pre-training phase, where BERT learns to predict masked portions of the input based on surrounding text. During the fine-tuning phase, the model is further trained using labeled data from the target task, and its parameters are jointly fine-tuned via supervised learning, allowing BERT to adapt its predictions to the specific task at hand. 

In contrast, prompt learning is a more recent paradigm that reformulates the NER task as textual prompts so that the model itself \textit{learns to predict the desired output}. Prompt learning has been shown to have better generalizability for unseen data with few or even no labeled samples \citep{agrawal2022large}. This is especially attractive for biomedical applications where annotations often require domain expertise and are not widely accessible due to data privacy. We compared BERT- and GPT-based models within the fine-tuning and prompt learning paradigms, respectively, due to their promising empirical performance on NER tasks in the biomedical domain \citep{yan2021unified, chen2021lightner, chen2023large}.
\subsubsection{Data Pre-processing and Fine-tuning BioClinicalBERT}
To pre-process our data, we split the texts into individual words (or subwords) with the BERT tokenizer and added special tokens (i.e., \texttt{CLS} and \texttt{SEP}) to the beginning and end of each tokenized sequence, respectively. We converted the tokens to their respective IDs, padded (or truncated) text sequences to obtain fixed-length inputs, and created an attention mask to distinguish between actual and padding tokens. Last, we mapped our labels, \{rare disease, disease, symptom, sign\}, to corresponding numerical values. 

We partitioned the data into a training, validation, and test set based on an 8:1:1 ratio. For the base architecture, we selected BioClinicalBERT \citep{alsentzer2019publicly}, a variant of BERT that was pre-trained on large-scale biomedical (PubMed, ClinicalTrials.gov) and clinical corpora (MIMIC-III \citep{johnson2016mimic}). To fine-tune BioClinicalBERT on our corpus, we trained the model on the training set and selected hyperparameters using the validation set. We used the test set to evaluate model performance.

\subsubsection{Prompt Learning using ChatGPT (GPT-3.5-turbo)}
In this section, we describe our approach to reformulating NER as a text generation task in the zero- and few-shot settings. The former refers to instructing the model to extract entities directly from an input text in the test set, and the latter is similar except we also provide an example of extracted entities from a training text.

\begin{table}
\resizebox{\textwidth}{!}{
\begin{tabular}{llll}
\hline
\hline
Setting   & Type & Prompt & Example\\
\hline
Zero-shot &   Simple   &  \texttt{\colorbox{pink}{Extract the exact names of [\textcolor{purple}{entity}]}},     & \texttt{\colorbox{pink}{Extract the exact names of rare diseases}, \colorbox{LightBlue1}{which are diseases}}\\
     &     &  \texttt{\colorbox{LightBlue1}{which are [\textcolor{purple}{defn}]}}, \texttt{from this passage} & \texttt{\colorbox{LightBlue1}{that affect a small number of individuals,} from this passage}         \\
      && \texttt{\colorbox{grannysmithapple}{and output them in a list:}} & \texttt{\colorbox{grannysmithapple}{and output them in a list:} "The exact prevalence and incidence} \\
    & & \texttt{"[\textcolor{purple}{text from test set}]".}&      \texttt{abetalipoproteinemia is unknown, but it is estimated to affect $\cdots$}\\
            &&& \hspace{18em} $\vdots$\\
          & & & \texttt{$\cdots$ incidence of consanguineous marriages. Symptoms usually }\\
&&& \texttt{become apparent during infancy."}\\
  &&&\\
  \cline{2-4}
          & Structured & \texttt{\#\#\#Task:} & \texttt{\#\#\#Task:}\\
          & & \texttt{\colorbox{pink}{Extract the exact names of [\textcolor{purple}{entity}]}} &\texttt{\colorbox{pink}{Extract the exact names of rare diseases}}\\
 & & \texttt{\colorbox{pink}{from the input text} and \colorbox{grannysmithapple}{output them}} & \texttt{\colorbox{pink}{from the input text} and \colorbox{grannysmithapple}{output them}} \\
  & & \texttt{\colorbox{grannysmithapple}{in a list.}} &\texttt{\colorbox{grannysmithapple}{in a list.}} \\
  &&&\\
   & & \texttt{\#\#\# Definition:} & \texttt{\#\#\# Definition:}\\
      & & \texttt{\colorbox{LightBlue1}{[\textcolor{purple}{entity}]s are defined as [\textcolor{purple}{defn}]}}. & \texttt{\colorbox{LightBlue1}{Rare diseases are defined as diseases that affect a small number}} \\
        &&& \texttt{\colorbox{LightBlue1}{of individuals}.}\\
        &&& \\
       & & \texttt{\#\#\# Input text: [\textcolor{purple}{text from test set}]}. &\texttt{\#\#\# Input text: "The exact prevalence and incidence of} \\
         &&& \texttt{abetalipoproteinemia is unknown, but it is estimated to affect $\cdots$}\\
            &&& \hspace{18em} $\vdots$\\
          & & & \texttt{$\cdots$ incidence of consanguineous marriages. Symptoms usually}\\
&&& \texttt{become apparent during infancy."}\\

           && \texttt{\#\#\# \colorbox{languidlavender}{Output:}}&\texttt{\#\#\# \colorbox{languidlavender}{Output:}}\\
             &&&\\
             \hline
Few-shot & Simple & \texttt{Passage: [\textcolor{purple}{text from training set}].}& \texttt{Passage: "Binder type nasomaxillary dysplasia is a rare congenital}\\
&& &\texttt{condition that affects males and females in equal numbers $\cdots$}\\
 &&& \hspace{18em} $\vdots$\\
 &&& \texttt{$\cdots$ suggests that Binder syndrome occurs in less than 1 per 10,000} \\
 &&& \texttt{live births."}\\
&& \texttt{\colorbox{Khaki1}{Extract the exact names of [\textcolor{purple}{entity}],}} & \texttt{\colorbox{Khaki1}{Extract the exact names of rare diseases, which are diseases}}\\
&& \texttt{\colorbox{Khaki1}{which are [\textcolor{purple}{defn}], from this passage}} & \texttt{\colorbox{Khaki1}{that affect a small number of individuals, from this passage}} \\
&& \texttt{\colorbox{Khaki1}{and output them in a list: }} &\texttt{\colorbox{Khaki1}{and output them in a list: }}\\
&& \texttt{\colorbox{Khaki1}{[\textcolor{purple}{gold standard training labels}].}}& \texttt{\colorbox{Khaki1}{Blinder type nasomaxillary dysplasia, Blinder syndrome}}\\
 &&& \\
  &  & \texttt{Passage: [\textcolor{purple}{text from test set}].}& \texttt{Passage: "The exact prevalence and incidence of } \\
  &&& \texttt{abetalipoproteinemia is unknown, but it is estimated to affect $\cdots$}\\
    &&& \hspace{18em} $\vdots$\\
    &&& \texttt{$\cdots$ incidence of consanguineous marriage. Symptoms usually}\\
    &&& \texttt{become apparent during infancy."}\\
&& \texttt{\colorbox{pink}{Extract the exact names of [\textcolor{purple}{entity}],}} &\texttt{\colorbox{pink}{Extract the exact names of [\textcolor{purple}{entity}],}}\\
&& \texttt{\colorbox{LightBlue1}{which are [\textcolor{purple}{defn}],} from this passage} &\texttt{\colorbox{LightBlue1}{which are [\textcolor{purple}{defn}],} from this passage}\\
&& \texttt{\colorbox{grannysmithapple}{and output them in a list:}} &\texttt{\colorbox{grannysmithapple}{and output them in a list:}}\\
\cline{2-4}
& Structured & \texttt{\#\#\# Task:} &\texttt{\#\#\# Task:} \\
& & \texttt{\colorbox{pink}{Extract the exact names of [\textcolor{purple}{entity}],}} &\texttt{\colorbox{pink}{Extract the exact names of rare diseases,}} \\
& &  \texttt{\colorbox{pink}{from the input text} and \colorbox{grannysmithapple}{output them}} & \texttt{\colorbox{pink}{from the input text} and \colorbox{grannysmithapple}{output them}}\\
& &  \texttt{\colorbox{grannysmithapple}{in a list.}} & \texttt{\colorbox{grannysmithapple}{in a list.}}\\
&&&\\
&& \texttt{\#\#\# Definition: } & \texttt{\#\#\# Definition: }\\
&&\texttt{\colorbox{LightBlue1}{[\textcolor{purple}{entity}]s are defined as [\textcolor{purple}{defn}].}} &\texttt{\colorbox{LightBlue1}{Rare diseases are defined as diseases that affect a small number}}\\
&&& \texttt{\colorbox{LightBlue1}{of individuals.}}\\
&&& \\
&& \texttt{\#\#\# Input text: [\textcolor{purple}{text from training set}]} & \texttt{\#\#\# Input text: "Blinder type nasomaxillary dysplasia is a rare}  \\
&&& \texttt{congenital condition that affects males and females in equal $\cdots$}\\
            &&& \hspace{18em} $\vdots$\\
        &&&\texttt{$\cdots$ suggests that Binder syndrome occurs in less than 1 per 10,000} \\
        &&& \texttt{live births."} \\
        &&&\\
&& \texttt{\#\#\# \colorbox{Khaki1}{Output: [\textcolor{purple}{gold standard training labels}]}} &\texttt{\#\#\# \colorbox{Khaki1}{Output: Blinder type nasomaxillary dysplasia, Blinder syndrome}} \\
        &&&\\
&& \texttt{\#\#\# Input text: [\textcolor{purple}{text from test set}]} & \texttt{\#\#\# Input text: "The exact prevalence and incidence of }\\
&& &\texttt{abetalipoproteinemia is unknown, but it is estimated to affect $\cdots$}\\ 
         &&& \hspace{18em} $\vdots$\\
          & & & \texttt{$\cdots$ incidence of consanguineous marriages. Symptoms usually }\\
&&& \texttt{become apparent during infancy."}\\
&&& \\
&& \texttt{\#\#\# \colorbox{languidlavender}{Output:} } & \texttt{\#\#\# \colorbox{languidlavender}{Output:} }\\
\hline
\end{tabular}}
\caption{Summary of prompts. Different parts of the prompt are color-coded as follows: \colorbox{pink}{Task instruction}, \colorbox{LightBlue1}{Task guidance}, \colorbox{grannysmithapple}{Output specification}, \colorbox{languidlavender}{Output retrieval}, and
\colorbox{Khaki1}{Specific example}.} 
\label{tab:prompts}
\end{table}

\textit{Prompt design}. Table~\ref{tab:prompts} provides a summary of prompts in the zero- and few-shot settings. The five main building blocks of our prompt designs were 1) task instruction, 2) task guidance, 3) output specification, 4) output retrieval, and, in the few-shot setting, 5) a specific example. Task instruction conveys the overall set of directions for NER in a specific but concise manner. To prevent ChatGPT from rephrasing entities, we instructed it to extract their \textit{exact} names from the input text. Task guidance provides entity definitions from the original RareDis annotation guidelines. The objective is to help ChatGPT differentiate between entity types within the context of the input text, as all four entities overlap semantically. Output specification instructs ChatGPT to output the extracted entities in a specific format to reduce post-processing workload. Output retrieval prompts the model to generate a response. In the few-shot setting, we also provided an example with an input text from the training set and its gold standard labels (i.e., entities labeled by the annotators).

\textit{Prompt format}. In each setting, we experimented with two prompt formats: simple and structured (Table~\ref{tab:prompts}). The former presents the prompt as a simple sentence, and the latter a structured list. The simple sentence is shorter in length and resembles human instructions provided in a conversational setting where different building blocks (i.e., task instruction, task guidance, and output specification) are woven together as a single unit. \cite{agrawal2022large} and \cite{hu2023zero} used a similar approach to extract medications and clinical entities, respectively. In contrast, the structured list resembles a recipe or outline that consists of multiple sub-prompts in a specific order. \cite{chen2023large} used a similar format for evaluating ChatGPT and GPT-4's NER performance on benchmark datasets. 

\textit{Few-shot example selection.} We explored two strategies for selecting an example text in the few-shot setting. The first strategy involved randomly selecting a text from the training set, and the second involved selecting the training text that was most similar to the test text. The motivation for the second strategy is that different rare diseases may have similar etiology, course of progression, and symptoms/signs. For example, Creutzfeldt-Jakob disease and CARASIL (cerebral autosomal recessive arteriopathy with subcortical infarcts and leukoencephalopathy) are neurological conditions that share similar signs, including progressive deterioration of cognitive processes and memory. Thus, providing a training text (and the corresponding gold standard entities) that was most similar to the test text may improve ChatGPT's performance. For each input text from the test set, we selected the training text that had the highest similarity score based on \texttt{spaCy} pre-trained word embeddings \citep{spacy}. 

\subsection{Evaluation}
To evaluate model performance on the test set, we computed the following evaluation metrics: precision, recall, and F1-score. Precision is the percentage of extracted entities found by the model that were correct, and recall the percentage of gold standard entities extracted by the model. F1 accounts for both metrics by taking the harmonic mean of precision and recall. We calculated these metrics under two evaluation settings: exact and relaxed. For an exact match, the extracted and true entity must share the same text span (i.e., boundary) and entity type. For a relaxed match, the extracted and true entity must overlap in boundary and have the same entity type. To ensure that stop words did not influence the evaluation, we removed them from both the gold standard and model-extracted entities. 

\section{Results}
\subsection{Overall Results}
Table~\ref{tab:results} provides a summary of the model performance by entity type. Overall, BioClinicalBERT achieved an F1-score of 0.689 under relaxed match. In the zero-shot setting, ChatGPT achieved F1-scores of 0.472 and 0.407 with the simple sentence and structured list prompts, respectively. Performance generally improved in the few-shot setting with F1-scores of 0.591 and 0.469; choosing the training text based on a similarity score led to additional improvement, resulting in F1 scores of 0.610 and 0.544. For some entities, ChatGPT had similar or better performance than its supervised counterpart, achieving F1-scores of 0.776 (vs. 0.755) and 0.725 (vs. 0.704) for rare diseases and signs, respectively, in the few-shot setting. Compared to prompts written as a structured list, simple sentences generally achieved similar or better performance, suggesting that ChatGPT may be more receptive to conversational prompts. Moreover, simple sentences required fewer tokens and were preferred over structured lists from a cost perspective. In the few-shot setting, selecting a training example that was similar to the input text led to better performance than random selection. 

   \begin{table}
\resizebox{1\textwidth}{!}{\begin{tabular}{|rccr|ccc|ccc|}
\hline
     &   &  & & \multicolumn{3}{c|}{Exact}  & \multicolumn{3}{c|}{Relaxed} \\
             \hline
  Paradigm &  Model&Setting &  Entity         & Precision & Recall & F1    & Precision  & Recall & F1    \\
             \hline
Pre-train&  BioClinicalBERT &Supervised& Rare disease &  0.689    &  0.720  & 0.704  &   0.772    & 0.739  &  0.755 \\
+ Fine-tune & && Disease   &  0.494   & 0.488 &  0.491& 0.532      &  0.538 & 0.535  \\
& && Sign  &  0.561  & 0.516  & 0.538 &     0.676 &  0.735 & 0.704 \\
&& & Symptom   &   0.667   &  0.630  &0.648 &   0.704  &  0.745 & 0.724 \\
&& & Overall &  0.600   & 0.583   & 0.591& 0.681      & 0.698  & 0.689\\
\hline
Prompt & ChatGPT &Zero-shot & Rare disease & 0.559& 0.409& 0.472& 0.843 & 0.694& 0.761 \\
learning  & &(Simple sentence)& Disease &0.109 & 0.240& 0.150 &0.200&0.437&0.274\\
&& & Sign &0.269 &0.380 &0.315 & 0.537 & 0.751 & 0.627 \\
&& & Symptom &0.070 & 0.619 & 0.126 & 0.084 & 0.762 & 0.155\\
& && Overall &0.203&0.369 & 0.262 & 0.365 & 0.670 & 0.472\\
\cline{3-10}
&&Zero-shot  & Rare disease & 0.765&0.489 &0.597 & 0.887  &0.634 &0.740  \\
&&(Structured list) & Disease &0.184 &0.210 &0.196  &0.261&0.293&0.276\\
&& & Sign & 0.266& 0.324& 0.292& 0.448 &0.543  &0.491  \\
&& & Symptom &0.063 &0.69  &0.116  & 0.079 &0.857  &0.145 \\
&& & Overall &0.226& 0.359&0.277  &0.331  &0.528  &0.407 \\
\cline{3-10}
& &Few-shot & Rare disease & 0.719 & 0.441 & 0.547 & 0.937 & 0.634 & 0.756  \\
&&(Simple sentence  & Disease &0.211&0.210&0.210 & 0.287 & 0.287 & 0.287\\
&& + random example) & Sign & 0.457 & 0.409 & 0.432 & 0.721 & 0.671 & 0.695\\
&& & Symptom & 0.279&0.452& 0.345& 0.294 & 0.476 & 0.364\\

& && Overall & 0.423 & 0.376 & 0.398 & 0.616 & 0.568 & 0.591\\
\cline{3-10}
&&Few-shot  & Rare disease & 0.569 & 0.532 & 0.550 & 0.750 &0.758  &0.754   \\
&&(Structured list & Disease & 0.151 &0.341  &0.209  & 0.211 &0.467  &0.291   \\
& &+ random example)& Sign & 0.273 & 0.406 &0.327  &  0.478& 0.698 &0.567   \\
&& & Symptom & 0.094  & 0.714  &0.166  & 0.107 &0.810  &0.189   \\
&& & Overall & 0.237 &0.440  &0.308  &0.361  &0.668  &0.469   \\
\cline{3-10}
&& Few-shot & Rare disease & 0.818& 0.484 &0.608 & 0.967 &0.634 &0.766 \\
&& (Simple sentence & Disease & 0.206& 0.246&0.224 &0.286 &0.341 &0.311 \\
& & + similar example) & Sign &0.441 &0.444 &0.443 &0.720 &0.730 &0.725 \\
& & & Symptom &0.260 &0.310 &0.283 & 0.308 &0.381 &0.340 \\
& & & Overall & 0.422&0.403 &0.412 & 0.617& 0.603& 0.610\\
\cline{3-10}
&& Few-shot & Rare disease & 0.590& 0.565& 0.577& 0.762&0.790 &0.776 \\
&& (Structured list & Disease & 0.199 & 0.437&0.273 &0.297 &0.653 &0.408 \\
& & + similar example) & Sign & 0.337 &0.487 &0.398 &0.561 &0.802 &0.660 \\
& & & Symptom & 0.093& 0.690& 0.164&0.114 &0.833 &0.200 \\
& & & Overall & 0.278& 0.506& 0.359&0.421 &0.769 &0.544 \\
\hline
\end{tabular}}
\caption{Summary of model performance by entity type.}\label{tab:results}
\end{table}

Among the four entities, rare diseases were associated with the highest accuracy for both models across all settings. In contrast, diseases were more challenging for both models. While BioClinicalBERT performed similarly at extracting signs and symptoms, ChatGPT achieved significantly better performance for signs. Because the only difference between the prompts for these entities was the task guidance, i.e., specifying symptoms as problems that \textit{cannot} be measured, whereas signs \textit{can} be measured, this finding suggests that ChatGPT is sensitive to even small variations in the prompt. 

\subsection{Detailed Error Analysis}
We conducted an in-depth error analysis to elucidate ChatGPT's performance. This analysis was crucial for gaining additional insight, as unlike other biomedical corpora, RareDis contains entities with overlapping semantics. Specifically, rare diseases are similar to diseases, and symptoms to signs. Depending on the context of the input text, diseases can also be symptoms or signs. 

In our analysis, we considered five types of errors: 1) incorrect boundary, 2) incorrect entity type, 3) incorrect boundary and entity type, 4) spurious, and 5) missed. The first and second refer to an extracted entity whose boundaries or type do not match those of the gold standard label, respectively. The third refers to the case where neither the extracted entity's boundaries nor type match those of the true label. Spurious entities are extracted entities that do not correspond to gold standard labels (false positive), and missed entities are entities that the model failed to extract (false negative). 

Table~\ref{tab:error} shows the distribution of errors in the few-shot setting under exact match. The most common error type for rare diseases is false negative (45\%) followed by incorrect entity type (31\%). In the case of entity type errors, ChatGPT tended to label rare diseases as diseases. These errors may be attributed to the fact that there is no single definition of rare diseases; rather, the definition can vary by country or location (i.e.,  a disease is a rare disease if it affects $< 200,000$ people in the United States or no more than 1 in 2,000 in the European Union). Moreover, this definition is subject to change over time, as a disease that used to be rare at the time of annotation may have become more prevalent, or vice versa. Because annotations are subjective, it's possible that what the domain experts deemed as rare diseases may not be reflected in textual information on the Internet before September 2021, ChatGPT's knowledge cut-off date. For instance, the annotators labeled ``gastrointestinal anthrax" and ``cutaneous anthrax" as rare diseases based on domain knowledge, but neither were listed in rare disease databases at the time of writing this manuscript. For diseases, signs, and symptoms, false positives and false negatives were the most common error types. Based on manual review, many of these errors can be attributed to the challenge of differentiating amongst these entities. Specifically, diseases could be signs or symptoms, and the difference between signs and symptoms is very subtle. In some cases, gold standard labels deviated from the definitions provided in the annotation guidelines, as the lack of abnormalities was also labeled as an entity (i.e., ``asymptomatic during infancy or childhood" was labeled as a symptom by the annotators). As such, a portion of false negatives could be attributed to these edge cases.    

\begin{table}
\resizebox{1\textwidth}{!}{\begin{tabular}{r|rrrrr|r}
\hline
\hline
         & Boundary \xmark  & Boundary \checkmark & Boundary   \xmark   & Spurious  & Missed & Total    \\
           & Entity type \checkmark & Entity type \xmark  & Entity type  \xmark     &  (False Pos.) & (False Neg.) & errors    \\
             \hline
Rare disease &  16 (10\%)   & 48 (31\%)   & 17 (11\%)  &     4 (3\%)    &  72 (45\%)  &   157 (100\%) \\
  & & & & & & \\
 \hline
Disease      &   11 (4\%)   & 7 (2\%) & 9 (3\%) &  147 (51\%)   &  116 (40\%) &  290 (100\%) \\
 & & & & & & \\
\hline
 Sign      &  64 (17\%)   & 8 (2\%)  & 5 (1\%) &   146 (40\%) &  148 (40\%) &   371 (100\%) \\
 & & & & & & \\
\hline
 Symptom      &  3 (4\%)    &  12 (16\%)   & 2 (3\%) &  34 (44\%)   &   25 (33\%) &    76 (100\%)\\
 & & & & & & \\
\hline
\end{tabular}}
\caption{Error analysis for ChatGPT in the few-shot setting under exact match.} \label{tab:error}
\end{table}

\section{Discussion}
In this work, we reformulated NER as a text generation task and established a benchmark for ChatGPT's performance on extracting rare disease phenotypes. Overall, while fine-tuning a pre-trained biomedical language model led to better performance, prompt learning with ChatGPT achieved similar or higher accuracy for some entities (i.e., rare diseases and signs) with a single example, demonstrating its potential for out-of-the-box NER in the few-shot setting. Given ChatGPT's performance in the zero-shot setting, the model could be leveraged as a pre-annotation tool to accelerate annotation start-up times for rare diseases and signs (F1-scores of 0.761 and 0.627, respectively). Overall, we recommend simple, sentence-based prompts, as they performed similarly or better than lists and were shorter in length, leading to lower computational cost.

While other studies explored supervised deep learning techniques for extracting rare disease phenotypes, ours is the first to study ChatGPT in the zero- and few-shot settings. \cite{segura2022exploring} compared the NER performance of base BERT, BioBERT, and ClinicalBERT, and found that ClinicalBERT had the highest overall F1-score (0.695). This is comparable to BioClinicalBERT's performance in the current study (0.689). \cite{fabregat2018deep} used support vector machines and neural networks with a long short-term memory architecture to extract disabilities associated with rare diseases and obtained an F1-score of 0.81. While this is much higher than the overall F1-score in the current study, \cite{fabregat2018deep} focused on extracting a single entity, i.e., disabilities, whereas our goal was to recognize and differentiate among four entities with overlapping semantics. \cite{hu2023zero} and \cite{chen2023large} evaluated ChatGPT on biomedical NER and found that it had lower performance than fine-tuning pre-trained language models. While our overall results aligned with this finding, we discovered that ChatGPT had similar or better performance on specific entities, suggesting that with appropriate prompt engineering, the model has the potential to match or outperform fine-tuned language models for certain entity types.

Our work has several potential limitations and extensions. First, we only had access to a subset of the RareDis corpus (832 out of 1041 texts), so our results may not fully reflect ChatGPT's performance across the entire spectrum of rare diseases. Second, the current work focuses on ChatGPT and does not include GPT-4 or other variants (e.g., LLaMA, Alpaca, etc.), so broadening the current set of experiments to include other large language models is a natural extension. Last, though manually created prompts are intuitive and interpretable, evidence suggests that small changes can lead to variations in performance \citep{cui2021template}. A promising alternative is to automate the prompt engineering process. To this end, \cite{gutierrez2022thinking} employed a semi-automated approach combining manually-created prompts with an automatic procedure to choose the best prompt combination with cross validation. In addition, fully-automated prompt learning approaches where the prompt is described directly in the embedding space of the underlying language model are also interesting extensions of the current work \citep{ma2021template, taylor2022clinical}.

The advent of large language models is creating unprecedented opportunities for rare disease phenotyping by automatically identifying and extracting diseases related concepts. While these models provide valuable insights and assistance, researchers and clinicians should critically evaluate model outputs and be well-informed of their limitations when considering them as tools for supporting rare disease diagnosis and treatment.

\section{Data and Code}
The RareDis corpus can be found using the link provided in \cite{martinez2022raredis}. The code for the current study can be found at \\\texttt{https://github.com/cathyshyr/rare\textunderscore disease\textunderscore phenotype\textunderscore extraction}. 

\clearpage

\bibliography{bibliography}  

\begin{thebibliography}{41}
\providecommand{\natexlab}[1]{#1}
\providecommand{\url}[1]{\texttt{#1}}
\expandafter\ifx\csname urlstyle\endcsname\relax
  \providecommand{\doi}[1]{doi: #1}\else
  \providecommand{\doi}{doi: \begingroup \urlstyle{rm}\Url}\fi

\bibitem[Agrawal et~al.(2022)Agrawal, Hegselmann, Lang, Kim, and
  Sontag]{agrawal2022large}
Monica Agrawal, Stefan Hegselmann, Hunter Lang, Yoon Kim, and David Sontag.
\newblock Large language models are few-shot clinical information extractors.
\newblock In \emph{Proceedings of the 2022 Conference on Empirical Methods in
  Natural Language Processing}, pages 1998--2022, 2022.

\bibitem[Ahmad et~al.(2020)Ahmad, Ricket, Hammill, Eskenazi, Robertson, Curtis,
  Dobi, Girotra, Haynes, Kizer, et~al.]{ahmad2020computable}
Faraz~S Ahmad, Iben~M Ricket, Bradley~G Hammill, Lisa Eskenazi, Holly~R
  Robertson, Lesley~H Curtis, Cecilia~D Dobi, Saket Girotra, Kevin Haynes,
  Jorge~R Kizer, et~al.
\newblock Computable phenotype implementation for a national, multicenter
  pragmatic clinical trial: lessons learned from adaptable.
\newblock \emph{Circulation: Cardiovascular Quality and Outcomes}, 13\penalty0
  (6):\penalty0 e006292, 2020.

\bibitem[Alsentzer et~al.(2019)Alsentzer, Murphy, Boag, Weng, Jin, Naumann, and
  McDermott]{alsentzer2019publicly}
Emily Alsentzer, John~R Murphy, Willie Boag, Wei-Hung Weng, Di~Jin, Tristan
  Naumann, and Matthew McDermott.
\newblock Publicly available clinical bert embeddings.
\newblock \emph{arXiv preprint arXiv:1904.03323}, 2019.

\bibitem[Carmichael et~al.(2015)Carmichael, Tsipis, Windmueller, Mandel, and
  Estrella]{carmichael2015going}
Nikkola Carmichael, Judith Tsipis, Gail Windmueller, Leslie Mandel, and Elicia
  Estrella.
\newblock “is it going to hurt?”: the impact of the diagnostic odyssey on
  children and their families.
\newblock \emph{Journal of Genetic Counseling}, 24:\penalty0 325--335, 2015.

\bibitem[Chapman et~al.(2021)Chapman, Dom{\'\i}nguez, Fairweather, Delaney, and
  Curcin]{chapman2021using}
Martin Chapman, Jes{\'u}s Dom{\'\i}nguez, Elliot Fairweather, Brendan Delaney,
  and Vasa Curcin.
\newblock Using computable phenotypes in point-of-care clinical trial
  recruitment.
\newblock In \emph{Public Health and Informatics-Proceedings of MIE 2021:
  Studies in Health Technology and Informatics}, pages 560--564. IOS Press,
  2021.

\bibitem[Chen et~al.(2023)Chen, Du, Hu, Keloth, Peng, Raja, Zhang, Lu, and
  Xu]{chen2023large}
Qingyu Chen, Jingcheng Du, Yan Hu, Vipina~Kuttichi Keloth, Xueqing Peng,
  Kalpana Raja, Rui Zhang, Zhiyong Lu, and Hua Xu.
\newblock Large language models in biomedical natural language processing:
  benchmarks, baselines, and recommendations.
\newblock \emph{arXiv preprint arXiv:2305.16326}, 2023.

\bibitem[Chen et~al.(2021)Chen, Li, Deng, Tan, Xu, Huang, Si, Chen, and
  Zhang]{chen2021lightner}
Xiang Chen, Lei Li, Shumin Deng, Chuanqi Tan, Changliang Xu, Fei Huang, Luo Si,
  Huajun Chen, and Ningyu Zhang.
\newblock Lightner: a lightweight tuning paradigm for low-resource ner via
  pluggable prompting.
\newblock \emph{arXiv preprint arXiv:2109.00720}, 2021.

\bibitem[Childerhose et~al.(2021)Childerhose, Rich, East, Kelley, Simmons,
  Finnila, Bowling, Amaral, Hiatt, Thompson,
  et~al.]{childerhose2021therapeutic}
Janet~Elizabeth Childerhose, Carla Rich, Kelly~M East, Whitley~V Kelley,
  Shirley Simmons, Candice~R Finnila, Kevin Bowling, Michelle Amaral, Susan~M
  Hiatt, Michelle Thompson, et~al.
\newblock The therapeutic odyssey: Positioning genomic sequencing in the search
  for a child’s best possible life.
\newblock \emph{AJOB Empirical Bioethics}, 12\penalty0 (3):\penalty0 179--189,
  2021.

\bibitem[Chung et~al.(2022)Chung, Project, Chu, and Chung]{chung2022rare}
Claudia Ching~Yan Chung, Hong Kong~Genome Project, Annie Tsz~Wai Chu, and Brian
  Hon~Yin Chung.
\newblock Rare disease emerging as a global public health priority.
\newblock \emph{Frontiers in public health}, 10:\penalty0 1028545, 2022.

\bibitem[Cohen and Biesecker(2010)]{cohen2010quality}
Julie~S Cohen and Barbara~B Biesecker.
\newblock Quality of life in rare genetic conditions: a systematic review of
  the literature.
\newblock \emph{American Journal of Medical Genetics Part A}, 152\penalty0
  (5):\penalty0 1136--1156, 2010.

\bibitem[Cui et~al.(2021)Cui, Wu, Liu, Yang, and Zhang]{cui2021template}
Leyang Cui, Yu~Wu, Jian Liu, Sen Yang, and Yue Zhang.
\newblock Template-based named entity recognition using bart.
\newblock \emph{arXiv preprint arXiv:2106.01760}, 2021.

\bibitem[Davis et~al.(2013)Davis, Sriram, Bush, Denny, and
  Haines]{davis2013automated}
Mary~F Davis, Subramaniam Sriram, William~S Bush, Joshua~C Denny, and
  Jonathan~L Haines.
\newblock Automated extraction of clinical traits of multiple sclerosis in
  electronic medical records.
\newblock \emph{Journal of the American Medical Informatics Association},
  20\penalty0 (e2):\penalty0 e334--e340, 2013.

\bibitem[Devlin et~al.(2018)Devlin, Chang, Lee, and Toutanova]{devlin2018bert}
Jacob Devlin, Ming-Wei Chang, Kenton Lee, and Kristina Toutanova.
\newblock Bert: Pre-training of deep bidirectional transformers for language
  understanding.
\newblock \emph{arXiv preprint arXiv:1810.04805}, 2018.

\bibitem[Fabregat et~al.(2018)Fabregat, Araujo, and
  Martinez-Romo]{fabregat2018deep}
Hermenegildo Fabregat, Lourdes Araujo, and Juan Martinez-Romo.
\newblock Deep neural models for extracting entities and relationships in the
  new rdd corpus relating disabilities and rare diseases.
\newblock \emph{Computer methods and programs in biomedicine}, 164:\penalty0
  121--129, 2018.

\bibitem[Guti{\'e}rrez et~al.(2022)Guti{\'e}rrez, McNeal, Washington, Chen, Li,
  Sun, and Su]{gutierrez2022thinking}
Bernal~Jim{\'e}nez Guti{\'e}rrez, Nikolas McNeal, Clay Washington, You Chen,
  Lang Li, Huan Sun, and Yu~Su.
\newblock Thinking about gpt-3 in-context learning for biomedical ie? think
  again.
\newblock \emph{arXiv preprint arXiv:2203.08410}, 2022.

\bibitem[Hu et~al.(2023)Hu, Ameer, Zuo, Peng, Zhou, Li, Li, Li, Jiang, and
  Xu]{hu2023zero}
Yan Hu, Iqra Ameer, Xu~Zuo, Xueqing Peng, Yujia Zhou, Zehan Li, Yiming Li,
  Jianfu Li, Xiaoqian Jiang, and Hua Xu.
\newblock Zero-shot clinical entity recognition using chatgpt.
\newblock \emph{arXiv preprint arXiv:2303.16416}, 2023.

\bibitem[Insights(2020)]{insights2020barriers}
NORD~Rare Insights.
\newblock Barriers to rare disease diagnosis, care and treatment in the us: a
  30-year comparative analysis, 2020.

\bibitem[Johnson et~al.(2016)Johnson, Pollard, Shen, Lehman, Feng, Ghassemi,
  Moody, Szolovits, Anthony~Celi, and Mark]{johnson2016mimic}
Alistair~EW Johnson, Tom~J Pollard, Lu~Shen, Li-wei~H Lehman, Mengling Feng,
  Mohammad Ghassemi, Benjamin Moody, Peter Szolovits, Leo Anthony~Celi, and
  Roger~G Mark.
\newblock Mimic-iii, a freely accessible critical care database.
\newblock \emph{Scientific data}, 3\penalty0 (1):\penalty0 1--9, 2016.

\bibitem[Lee et~al.(2023)Lee, Goldberg, and Kohane]{lee2023ai}
Peter Lee, Carey Goldberg, and Isaac Kohane.
\newblock The ai revolution in medicine: Gpt-4 and beyond, 2023.

\bibitem[Li et~al.(2015)Li, Jin, Jiang, Song, and Huang]{li2015biomedical}
Lishuang Li, Liuke Jin, Zhenchao Jiang, Dingxin Song, and Degen Huang.
\newblock Biomedical named entity recognition based on extended recurrent
  neural networks.
\newblock In \emph{2015 IEEE International Conference on bioinformatics and
  biomedicine (BIBM)}, pages 649--652. IEEE, 2015.

\bibitem[Liu et~al.(2023)Liu, Yuan, Fu, Jiang, Hayashi, and Neubig]{liu2023pre}
Pengfei Liu, Weizhe Yuan, Jinlan Fu, Zhengbao Jiang, Hiroaki Hayashi, and
  Graham Neubig.
\newblock Pre-train, prompt, and predict: A systematic survey of prompting
  methods in natural language processing.
\newblock \emph{ACM Computing Surveys}, 55\penalty0 (9):\penalty0 1--35, 2023.

\bibitem[Lo~Barco et~al.(2021)Lo~Barco, Kuchenbuch, Garcelon, Neuraz, and
  Nabbout]{lo2021improving}
Tommaso Lo~Barco, Mathieu Kuchenbuch, Nicolas Garcelon, Antoine Neuraz, and
  Rima Nabbout.
\newblock Improving early diagnosis of rare diseases using natural language
  processing in unstructured medical records: an illustration from dravet
  syndrome.
\newblock \emph{Orphanet Journal of Rare Diseases}, 16:\penalty0 1--12, 2021.

\bibitem[Ma et~al.(2021)Ma, Zhou, Gui, Tan, Li, Zhang, and
  Huang]{ma2021template}
Ruotian Ma, Xin Zhou, Tao Gui, Yiding Tan, Linyang Li, Qi~Zhang, and Xuanjing
  Huang.
\newblock Template-free prompt tuning for few-shot ner.
\newblock \emph{arXiv preprint arXiv:2109.13532}, 2021.

\bibitem[Macnamara et~al.(2019)Macnamara, D’Souza, Tifft,
  et~al.]{macnamara2019undiagnosed}
Ellen~F Macnamara, Precilla D’Souza, Cynthia~J Tifft, et~al.
\newblock The undiagnosed diseases program: Approach to diagnosis.
\newblock \emph{Translational Science of Rare Diseases}, 4\penalty0
  (3-4):\penalty0 179--188, 2019.

\bibitem[Mart{\'\i}nez-deMiguel et~al.(2022)Mart{\'\i}nez-deMiguel,
  Segura-Bedmar, Chac{\'o}n-Solano, and Guerrero-Aspizua]{martinez2022raredis}
Claudia Mart{\'\i}nez-deMiguel, Isabel Segura-Bedmar, Esteban
  Chac{\'o}n-Solano, and Sara Guerrero-Aspizua.
\newblock The raredis corpus: a corpus annotated with rare diseases, their
  signs and symptoms.
\newblock \emph{Journal of Biomedical Informatics}, 125:\penalty0 103961, 2022.

\bibitem[Mehnen et~al.(2023)Mehnen, Gruarin, Vasileva, and
  Knapp]{mehnen2023chatgpt}
Lars Mehnen, Stefanie Gruarin, Mina Vasileva, and Bernhard Knapp.
\newblock Chatgpt as a medical doctor? a diagnostic accuracy study on common
  and rare diseases.
\newblock \emph{medRxiv}, pages 2023--04, 2023.

\bibitem[Nguengang~Wakap et~al.(2020)Nguengang~Wakap, Lambert, Olry, Rodwell,
  Gueydan, Lanneau, Murphy, Le~Cam, and Rath]{nguengang2020estimating}
St{\'e}phanie Nguengang~Wakap, Deborah~M Lambert, Annie Olry, Charlotte
  Rodwell, Charlotte Gueydan, Val{\'e}rie Lanneau, Daniel Murphy, Yann Le~Cam,
  and Ana Rath.
\newblock Estimating cumulative point prevalence of rare diseases: analysis of
  the orphanet database.
\newblock \emph{European Journal of Human Genetics}, 28\penalty0 (2):\penalty0
  165--173, 2020.

\bibitem[Nigwekar et~al.(2014)Nigwekar, Solid, Ankers, Malhotra, Eggert,
  Turchin, Thadhani, and Herzog]{nigwekar2014quantifying}
Sagar~U Nigwekar, Craig~A Solid, Elizabeth Ankers, Rajeev Malhotra, William
  Eggert, Alexander Turchin, Ravi~I Thadhani, and Charles~A Herzog.
\newblock Quantifying a rare disease in administrative data: the example of
  calciphylaxis.
\newblock \emph{Journal of general internal medicine}, 29:\penalty0 724--731,
  2014.

\bibitem[OpenAI(2022)]{ChatGPT}
OpenAI.
\newblock Introducing chatgpt.
\newblock \url{https://openai.com/blog/chatgpt}, 2022.

\bibitem[Patil et~al.(2020)Patil, Patil, and Pawar]{patil2020named}
Nita Patil, Ajay Patil, and BV~Pawar.
\newblock Named entity recognition using conditional random fields.
\newblock \emph{Procedia Computer Science}, 167:\penalty0 1181--1188, 2020.

\bibitem[Radford et~al.(2018)Radford, Narasimhan, Salimans, Sutskever,
  et~al.]{radford2018improving}
Alec Radford, Karthik Narasimhan, Tim Salimans, Ilya Sutskever, et~al.
\newblock Improving language understanding by generative pre-training.
\newblock 2018.

\bibitem[Rath et~al.(2012)Rath, Olry, Dhombres, Brandt, Urbero, and
  Ayme]{rath2012representation}
Ana Rath, Annie Olry, Ferdinand Dhombres, Maja~Mili{\v{c}}i{\'c} Brandt, Bruno
  Urbero, and Segolene Ayme.
\newblock Representation of rare diseases in health information systems: the
  orphanet approach to serve a wide range of end users.
\newblock \emph{Human mutation}, 33\penalty0 (5):\penalty0 803--808, 2012.

\bibitem[Segura-Bedmar et~al.(2022)Segura-Bedmar, Camino-Perdones, and
  Guerrero-Aspizua]{segura2022exploring}
Isabel Segura-Bedmar, David Camino-Perdones, and Sara Guerrero-Aspizua.
\newblock Exploring deep learning methods for recognizing rare diseases and
  their clinical manifestations from texts.
\newblock \emph{BMC bioinformatics}, 23\penalty0 (1):\penalty0 263, 2022.

\bibitem[spaCy()]{spacy}
spaCy.
\newblock Industrial-strength natural language processing in python.
\newblock \url{https://spacy.io}.

\bibitem[Stenetorp et~al.(2012)Stenetorp, Pyysalo, Topi{\'c}, Ohta, Ananiadou,
  and Tsujii]{stenetorp2012brat}
Pontus Stenetorp, Sampo Pyysalo, Goran Topi{\'c}, Tomoko Ohta, Sophia
  Ananiadou, and Jun’ichi Tsujii.
\newblock Brat: a web-based tool for nlp-assisted text annotation.
\newblock In \emph{Proceedings of the Demonstrations at the 13th Conference of
  the European Chapter of the Association for Computational Linguistics}, pages
  102--107, 2012.

\bibitem[Taylor et~al.(2022)Taylor, Zhang, Joyce, Nevado-Holgado, and
  Kormilitzin]{taylor2022clinical}
Niall Taylor, Yi~Zhang, Dan Joyce, Alejo Nevado-Holgado, and Andrey
  Kormilitzin.
\newblock Clinical prompt learning with frozen language models.
\newblock \emph{arXiv preprint arXiv:2205.05535}, 2022.

\bibitem[Tifft and Adams(2014)]{tifft2014national}
Cynthia~J Tifft and David~R Adams.
\newblock The national institutes of health undiagnosed diseases program.
\newblock \emph{Current opinion in pediatrics}, 26\penalty0 (6):\penalty0 626,
  2014.

\bibitem[Vaswani et~al.(2017)Vaswani, Shazeer, Parmar, Uszkoreit, Jones, Gomez,
  Kaiser, and Polosukhin]{vaswani2017attention}
Ashish Vaswani, Noam Shazeer, Niki Parmar, Jakob Uszkoreit, Llion Jones,
  Aidan~N Gomez, {\L}ukasz Kaiser, and Illia Polosukhin.
\newblock Attention is all you need.
\newblock \emph{Advances in neural information processing systems}, 30, 2017.

\bibitem[Wang et~al.(2018)Wang, Wang, Rastegar-Mojarad, Moon, Shen, Afzal, Liu,
  Zeng, Mehrabi, Sohn, et~al.]{wang2018clinical}
Yanshan Wang, Liwei Wang, Majid Rastegar-Mojarad, Sungrim Moon, Feichen Shen,
  Naveed Afzal, Sijia Liu, Yuqun Zeng, Saeed Mehrabi, Sunghwan Sohn, et~al.
\newblock Clinical information extraction applications: a literature review.
\newblock \emph{Journal of biomedical informatics}, 77:\penalty0 34--49, 2018.

\bibitem[Yan et~al.(2021)Yan, Gui, Dai, Guo, Zhang, and Qiu]{yan2021unified}
Hang Yan, Tao Gui, Junqi Dai, Qipeng Guo, Zheng Zhang, and Xipeng Qiu.
\newblock A unified generative framework for various ner subtasks.
\newblock \emph{arXiv preprint arXiv:2106.01223}, 2021.

\bibitem[Yang et~al.(2022)Yang, Cintina, Pariser, Oehrlein, Sullivan, and
  Kennedy]{yang2022national}
Grace Yang, Inna Cintina, Anne Pariser, Elisabeth Oehrlein, Jamie Sullivan, and
  Annie Kennedy.
\newblock The national economic burden of rare disease in the united states in
  2019.
\newblock \emph{Orphanet journal of rare diseases}, 17\penalty0 (1):\penalty0
  1--11, 2022.

\end{thebibliography}
\end{document}